\documentclass[letterpaper, 10 pt, conference]{ieeeconf}  % Comment this line out if you need a4paper

\IEEEoverridecommandlockouts                              % This command is only needed if 
\usepackage{times}
\usepackage{epsfig}
\usepackage{graphicx}
\usepackage{amsmath}

\usepackage{amssymb}
\usepackage{graphics} % for pdf, bitmapped graphics files
\usepackage{epsfig} % for postscript graphics files\usepackagr{cite}
\usepackage{mathptmx} % assumes new font selection scheme installed
\usepackage{multicol}

\usepackage{url}
\usepackage{booktabs} % For prettier tables
\usepackage{multirow}% http://ctan.org/pkg/multirow
\usepackage{float}
\usepackage{placeins}
\usepackage{graphicx} 
\usepackage[font=small]{caption}
\usepackage{subcaption}

\usepackage{lmodern}
\usepackage{siunitx}
\usepackage{etoolbox}
\robustify\bfseries
\usepackage{multicol}

\usepackage{fixme}

\usepackage{array}

\usepackage[ruled, shortend, linesnumbered]{algorithm2e}
\usepackage[pagebackref=true,breaklinks=true,colorlinks,bookmarks=true]{hyperref}

\usepackage{breqn}  % to break long equation
\newcommand*\blue{\color{black}}

\title{\LARGE \bf Stein Particle Filter for Nonlinear, \\ Non-Gaussian State Estimation}

\author{Fahira Afzal Maken$^{1,*}$, Fabio Ramos$^{1,2}$ and Lionel Ott$^{3}$% <-this % stops a space
	\thanks{$^*$ Corresponding author: fafz3958@uni.sydney.edu.au}
	\thanks{$^1$ School of Computer Science, The University of Sydney, Australia}
	\thanks{$^2$ NVIDIA, USA}
	\thanks{$^3$ ETH Z\"urich, Switzerland}%
}

\begin{document}

\maketitle
\thispagestyle{empty}
\pagestyle{empty}

%%%%%%%%%%%%%%%%%%%%%%%%%%%%%%%%%%%%%%%%%%%%%%%%%%%%%%%%%%%%%%%%%%%%%%%%%%%%%%%%

\begin{abstract}
    Estimation of a dynamical system's latent state subject to sensor noise and model inaccuracies remains a critical yet difficult problem in robotics. While Kalman filters provide the optimal solution in the least squared sense for linear and Gaussian noise problems, the general nonlinear and non-Gaussian noise case is significantly more complicated, typically relying on sampling strategies that are limited to low-dimensional state spaces. In this paper we devise a general inference procedure for filtering of nonlinear, non-Gaussian dynamical systems that exploits the differentiability of both the update and prediction models to scale to higher dimensional spaces. Our method, Stein particle filter, can be seen as a deterministic flow of particles, embedded in a reproducing kernel Hilbert space, from an initial state to the desirable posterior. The particles evolve jointly to conform to a posterior approximation while interacting with each other through a repulsive force. We evaluate the method in simulation and in complex localization tasks while comparing it to sequential Monte Carlo solutions. 
    
\end{abstract}

% +-----------------------------------------------------------------------------
% | Introduction
% +-----------------------------------------------------------------------------
\section{Introduction}

State estimation is a core component of many robotic systems and is used in applications ranging from self-driving vehicles for tasks such as ego-vehicle state estimation and dynamic object tracking, to grasping and manipulation for precise object pose estimation. The most widely employed type of state estimation filters are Bayesian filters such as Kalman filters and non-parametric alternatives such as particle filters. A Bayesian filter determines the state of a system based on an {\em a-priori} specified process model and an update model. The posterior estimate over the system's state is computed via Bayes' rule combining a prior distribution over the system state and a likelihood function which captures the relation between the system state and observations.

A Kalman filter~\cite{kalman_1960} is optimal for linear systems with Gaussian noise while extensions to it such as the extended Kalman filter (EKF)~\cite{ekf_re} relax the linearity assumption by computing local linearizations to handle nonlinear systems. These approaches are widely used due to their computational efficiency and simplicity. However, the assumptions of a parametric representation limit their applicability in complex and multi-modal scenarios. In those scenarios a non-parametric particle filter~\cite{doucet_2001_introduction}, which represents the state using a collection of particles, or samples, rather than a parametric distribution, is often employed. The trade-off for this increased expressiveness comes in the form of computational complexity making particle filters challenging to use in high-dimensional problem domains due to the number of particles required growing exponentially.

In this paper, we propose a novel filter method based on a Quasi-Newton extension of Stein Variational Gradient Descent (SVGD)~\cite{steinNIPS2016_6338} to address the scalability issues of particle filters while remaining flexible to accommodate multi-modal posteriors. Similar to particle filters, SVGD approximates the posterior distribution using a set of particles. However, unlike a particle filter which resamples particles based on their weight, in SVGD particles are transported towards the posterior distribution along the update model's gradients. This optimization is embedded in a reproducing kernel Hilbert space (RKHS) which provides a closed-form expression of the posterior distribution's gradient. To ensure the full distribution is recovered as opposed to only the most likely mode the interaction between particles is taken into account via a smoothing kernel. The net result of this is that fewer particles are required to obtain a good approximation. Additionally, since SVGD transports particles towards the posterior without any resampling the issues arising from particle deprivation in standard particle filters are not present. {\blue A Python implementation of the proposed method is publicly available\footnote{\url{https://bitbucket.org/fafz/stein_particle_filter}}}.

\noindent\textbf{Contribution:} 
The main contribution of the paper is a gradient-based particle filtering algorithm that incorporates second-order information to scale the method to higher dimensional problems.
{\blue The proposed method exploits the differentiability of the update and process models together with the flexibility of SVGD to model complex dynamical systems. We incorporate second-order information with L-BFGS optimization~\cite{Nocedal2006Numerical} into the gradient flow of particles to alleviate problems faced by non-parametric filtering methods in higher dimensional problems.} Experiments for general state estimation and localization applications demonstrate the practical properties of the method.

% +-----------------------------------------------------------------------------
% | Related Work
% +-----------------------------------------------------------------------------
 \section{Related Work}

Filtering has long been used in robotics and other fields for a wide range of applications which needed to estimate parameters of some system based on observations. For many state estimation systems a Kalman filter \cite{kalman_1960} or extensions to it such as the extended Kalman filter~\cite{ekf_re} and unscented Kalman filter (UKF) \cite{ukf_rebut} are used. This includes real-time MAV state estimation \cite{weiss_2012}, self-driving vehicles \cite{thrun_2006_stanley}, and NASA's Ingenuity Mars helicopter \cite{grip_2019}. Despite their widespread use in robotics and elsewhere their restrictive assumptions of uni-modality and Gaussian noise makes them a bad fit for a variety of challenging tasks. This includes WiFi-based indoor localization \cite{kothari_2012} and complex sensor fusion setups \cite{thomas_2007} where multi-modality and non-Gaussianity are expected to occur.

Despite the more expressive nature of particle filters, when compared to Kalman filters, they have a major drawback, namely computational complexity. To ensure proper operation of a particle filter a sufficient number of particles has to be used which increases with the dimensionality of the problem. For a planar 3 DoF localization problem it is not uncommon to use 1000 particles or more to ensure good performance. Consequently, there is a wide range of methods that attempt to alleviate this challenge of particle filters, including methods such as Hybrid Monte Carlo (HMC) filters~\cite{duane_1987, choo_2001}, corrective gradient refinement filter (CGR)~\cite{biswas_2011}, Rao-Blackwellised particle filters (RBPF)~\cite{rao_black_2001}, and Gaussian particle filters (GPF)~\cite{gaussian_pf}. HMC filters uses a set of Markov chains to generate a fixed number of samples and explore the state space. This is combined with more advanced Markov Chain Monte Carlo (MCMC) techniques such as Hamiltonian dynamics and Metropolis rejection test to improve sample efficiency. Nonetheless, even if the number of particles is reduced the overall number of samples remains large in part due to the burn-in required by MCMC. CGR adds a refinement and acceptance step into the standard filter loop with the goal to redistribute samples to better capture uncertainty. To this end CGR uses the gradient of the update model to correct estimates of particles that disagree with the observation model. In contrast to our method, which also uses the gradient of the update model, CGR performs a resampling step. RBPF require structural knowledge of the posterior distribution and increases the particle efficiency by marginalizing out the tractable substructure of the filter from the posterior distribution. In RBPF each particle is equipped with a Kalman filter or a hidden Markov model filter, to perform marginalization, which needs to be updated in each iteration making RBPF computationally more expensive than a particle filter. GPF approximates the posterior distribution by single Gaussian and is sensitive to the linearization errors.

Particle filters also suffer from particle impoverishment~\cite{pf_tut} which causes all particles to collapse to a small subset of particles in the resampling step, thus reducing diversity. Different methods have been proposed to address this problem by either keeping track of hypothesis or redistributing particles. Clustered particle filters~\cite{cluster_pf} for example preserve the particles for multiple likely hypotheses. KLD adaptive sampling~\cite{kld_sampling} adapts the number of particles based on the uncertainty in the belief potentially maintaining the diversity. {\blue To overcome particle degenarcy issues \cite{rebut_uncent_pf} runs an individual UKF for each particle.} Sensor resetting~\cite{reset_loc} is an approach that initializes particles based on hypothesis from the observations when the state estimate is uncertain. However, sensor resetting depends on the likelihood of the current observations given the current state, which makes it sensitive to noise in the observations~\cite{sensor_reset_multi}. Our proposed method does not suffer from this problem as SVGD naturally prevents all particles from collapsing onto each other and forces particles to be distributed over the full posterior via the gradient information of the update model.

An emerging class of filtering methods attempts to combine the correctness of Monte Carlo (MC) sampling with the speed of variational inference (VI).  
{\blue The mapping particle filter (MPF)~\cite{mapping_particle_f}   
uses first order gradient information to represent the posterior distribution in a 
SVGD~\cite{steinNIPS2016_6338} framework, and hence can be slow to converge~\cite{var_red}. Moreover, MPF uses an isotropic kernel to weight and disperse the particles which can be problematic in capturing the structure of the posterior distribution. Our method generalizes MPF by incorporating second order information into the gradient flow as well as in transforming the kernel to distribute particles to high-probability regions. This results in an improved convergence rate for higher dimensional problems. }

% +-----------------------------------------------------------------------------
% | Preliminaries
% +-----------------------------------------------------------------------------
\section{Preliminaries}

We review the basics of Bayes filtering and Stein variational gradient descent~\cite{steinNIPS2016_6338} as a generic technique for Bayesian inference below. Our proposed solution to filtering is presented in section~\ref{sec:spf}.

\subsection{Bayes Filtering}
We consider a discrete-time hidden Markov model with latent states $X_t = \{\mathbf{x}_{1:t}\}$, observations $Z_t = \{\mathbf{z}_{1:t}\}$, and controls $U_t=\{\mathbf{u}_{1:t}\}$ indexed by the time sequence $1,2,\ldots,t$. The evolution of the state sequence is given by $\mathbf{x}_t = \mathbf{f}_{t}(\mathbf{x}_{t-1}, \mathbf{u}_{t-1}, \mathbf{v}_{t-1})$ where $\mathbf{f}_t: \mathbb{R}^{d_x} \times \mathbb{R}^{d_u} \times \mathbb{R}^{d_v} \rightarrow \mathbb{R}^{d_x}$ is potentially a nonlinear function of the state $\mathbf{x}_{t-1}$, control $\mathbf{u}_{t-1}$, and $\mathbf{v}_{t-1}$ are samples from an independent and identically distributed noise process $p(\mathbf{v}_{1:t})=\prod_{i=1}^t p(\mathbf{v}_i)$. In filtering, we are interested in recursively estimating $\mathbf{x}_t$ given measurements (or observations) $\mathbf{z}_t=\mathbf{h}(\mathbf{x}_t, \mathbf{n}_t)$, where $\mathbf{h}:\mathbb{R}^{d_x} \times \mathbb{R}^{d_n} \rightarrow \mathbb{R}^{d_z}$ can be a nonlinear sensor model with i.i.d measurement noise sequence $p(\mathbf{n}_{1:t})=\prod_{i=1}^t p(\mathbf{n}_i)$. The dimensions for $\mathbf{x}_t$, $\mathbf{u}_t$, $\mathbf{v}_v$, $\mathbf{z}_t$ and $\mathbf{n}_t$ are denoted by $d_x$, $d_u$, $d_v$, $d_z$ and $d_n$ respectively.  
Within a Bayesian framework, this problem amounts to computing a belief for the current
state $\mathbf{x}_t$, given by $bel(\mathbf{x}_t)= p(\mathbf{x}_t|\mathbf{z}_{1:t}, \mathbf{u}_{1:t})$, conditioned on the history of observations $Z_t$ and control actions $U_t$. Assuming an initial state distribution or prior $p(\mathbf{x}_0|\mathbf{z}_0)=p(\mathbf{x}_0)$, and using the Markov assumption, the posterior can be calculated iteratively following a two step procedure typically referred to as prediction and update. 

The prediction step uses the state transition model %or a motion model
$p(\mathbf{x}_t|\mathbf{x}_{t-1},\mathbf{u}_t)=p(\mathbf{x}_t|\mathbf{f}_{t}(\mathbf{x}_{t-1}, \mathbf{u}_{t}, \mathbf{v}_{t-1}))$ and the posterior of $p(\mathbf{x}_{t-1}|\mathbf{z}_{1:t-1}, \mathbf{u}_{1:t-1})$ obtained in the previous iteration to predict $p(\mathbf{x}_{t}|\mathbf{z}_{1:t-1}, \mathbf{u}_{1:t})$, following the Chapman-Kolmogorov equation:

\begin{equation}
\widetilde{bel}(\mathbf{x}_t)=\int p(\mathbf{x}_t|\mathbf{x}_{t-1},\mathbf{u}_{t})\;p(\mathbf{x}_{t-1}|\mathbf{z}_{1:t-1}, \mathbf{u}_{1:t-1})\;d\mathbf{x}_{t-1}.
\label{eq:predicted belief}
\end{equation}

The update step computes the likelihood of an observation ($\mathbf{z}_t$) given the state ($\mathbf{x}_t$), using the sensor model $p(\mathbf{z}_t|\mathbf{x}_t)=p(\mathbf{z}_t|\mathbf{h}(\mathbf{x}_t, \mathbf{n}_t))$, and updates the belief for the current step following Bayes' rule:

\begin{equation}
{bel}(\mathbf{x}_t)=\eta p(\mathbf{z}_t|\mathbf{x}_{t}) \;\widetilde{bel}(\mathbf{x}_{t-1}),
\label{eq:corrected_belief}
\end{equation}
where $\eta$ is a normalization constant given by
\vskip-5mm
\begin{equation}
   \eta^{-1} = p(\mathbf{z}_t|\mathbf{z}_{1:t-1})=\int p(\mathbf{z}_t|\mathbf{x}_t) p(\mathbf{x}_t|\mathbf{z}_{1:t-1}, \mathbf{u}_{1:t-1}) d\mathbf{x}_t.
\end{equation}
\vskip-7mm
\subsection{Particle Filters}

 As the integrals in \eqref{eq:predicted belief} and \eqref{eq:corrected_belief} are typically not tractable, particle filters~\cite{gordon_1993} approximate the belief using $N$ weighted particles, ${bel}(\mathbf{x}_t)=\{(\mathbf{x}_t^j,w_t^j)\}_{j=1}^{N}$, generated by Monte Carlo simulation. A particle filter updates the belief in a three step process of prediction, update and resampling. In the prediction step, a particle filter samples a motion model to move each particle stochastically in \eqref{eq:pf_pred}. In the update step, the update \eqref{eq:corrected_belief} is achieved in \eqref{eq:pf_correct} by assigning weights to each particle using an observation likelihood. 
 In the resampling step, these weighted particles are then resampled proportionally to their weights: 
 \iffalse
    \begin{align}
        \forall_j \; \; \mathbf{x}_t^j &\sim  \;p(\mathbf{x}_t^j|\mathbf{x}_{t-1}^j,\mathbf{u}_{t}), \label{eq:pf_pred} \\
        \forall_j \; \;w_t^j &= p(\mathbf{z}_t|\mathbf{x}_{t}^j).
        \label{eq:pf_correct}
    \end{align}
\fi
\noindent\begin{minipage}{.55\linewidth}
\begin{equation}
 \forall_j \; \; \mathbf{x}_t^j \sim  \;p(\mathbf{x}_t^j|\mathbf{x}_{t-1}^j,\mathbf{u}_{t}),
 \label{eq:pf_pred}
\end{equation}
\end{minipage}%
\begin{minipage}{.4\linewidth}
\begin{equation}
 \forall_j \; \;w_t^j = p(\mathbf{z}_t|\mathbf{x}_{t}^j).
  \label{eq:pf_correct}
\end{equation}
\end{minipage}
\subsection{Stein Variational Gradient Descent (SVGD)}
\label{svgd}

Similar to particle filters, SVGD approximates an intractable but differentiable posterior (target) distribution $p(\mathbf{x}|\mathbf{z})$ with a nonparametric distribution $q(\mathbf{x})=\frac{1}{N}\sum_{i=j}^N \delta(\mathbf{x}-\mathbf{x}^j)$ represented by a set of $N$ particles $\{\mathbf{x}^j\}_{j=1}^N$, where $\delta(\cdot)$ is the Dirac delta function. 
In contrast to particle filters, the particles are transported deterministically along the optimal gradient direction to match the posterior distribution in a series of steps, each minimizing the KL divergence between the true posterior $p(\mathbf{x}|\mathbf{z})$ and the variational approximation $q_{[\epsilon\boldsymbol{\phi}]}(\mathbf{x})$ as follows:
\vskip-8mm
\begin{align}
&T(\mathbf{x}^j_{k+1}) =   \mathbf{x}^j_k +  \epsilon \boldsymbol{\phi}^*_k (\mathbf{x}^j), ~ \forall j = 1, \ldots, N, \label{eq:phi_star_up}\\\text{ }
&\boldsymbol{\phi}_k^* =   \operatorname*{argmax}_{\boldsymbol{\phi} \in \mathcal{B}_k }  \bigg(  -   \frac{d}{d\epsilon} \text KL(q_{[\epsilon\boldsymbol{\phi}]} ~|| ~ p) \big |_{\epsilon = 0}  \bigg), 
\label{eq:phi_star}
\end{align}
where $\epsilon$ is a small step size, and 
$\boldsymbol{\phi}_k^* \colon \mathbb R^d \to \mathbb R^d$ denotes the optimal transport function representing the direction to move the approximation within the functional space $\mathcal{B}_k$ closer to the target. $q_{[\epsilon\boldsymbol{\phi}]}$ denotes the distribution of the updated particles, $\mathbf{x}_{k+1} = \mathbf{x}_k + \epsilon \boldsymbol{\phi}_k(\mathbf{x})$ which decreases the KL divergence along the steepest direction from $q$ by $\epsilon$. The iterative transformation in \eqref{eq:phi_star} yields $\boldsymbol{\phi}^*(x)=0$ when the KL divergence converges to a local minimum. At this stage, the final $q$ represents the nonparametric variational approximation to the target.

To obtain a closed-form solution, SVGD chooses $\mathcal{B}_k$ to be in the unit ball of a vector-valued reproducing kernel Hilbert space (RKHS), $\mathcal{H}_k^d = \mathcal{H}_k \times \cdots \times \mathcal{H}_k$,
where $\mathcal{H}_k$ is a RKHS formed by scalar-valued functions associated with a positive definite kernel $k(\mathbf{x},\mathbf{x}')$, that is, 
$\mathcal{B}_k = \{\boldsymbol{\phi} \in \mathcal{H}_k^d  \colon ||\boldsymbol{\phi}||_{\mathcal{H}_k^d}\leq  1 \}.$ The objective function \eqref{eq:phi_star} can be expressed as a linear functional of $\phi$ that connects to the Stein operator $\mathcal{A} \boldsymbol{\phi}(\mathbf{x})$~\cite{steinNIPS2016_6338},
\vskip-4mm
\begin{align}
&-\frac{d}{d\epsilon} \text KL(q_{[\epsilon\boldsymbol{\phi}]} ~|| ~ p) \big |_{\epsilon = 0}  = \mathbb E_{\mathbf{x}\sim q}[\text trace(\mathcal{A}  \boldsymbol{\phi}(\mathbf{x}))],
\\
 &\mathcal{A} \boldsymbol{\phi}(\mathbf{x})  = {\blue \boldsymbol{\phi} (\mathbf{x}) \nabla_{\mathbf{x}} \log p(\mathbf{x})^\top}  + \nabla_{\mathbf{x}}\boldsymbol{\phi}(\mathbf{x}), 
\label{eq:stein_operator}
\end{align}
where  $\nabla_{\mathbf{x}} \log p(\mathbf{x})$ is the gradient of the log posterior. 
Note that the Stein operator depends on the posterior distribution only through $\nabla_{\mathbf{x}} \log p(\mathbf{x})$, which does not require the normalization constant (also known as marginal likelihood) which is generally intractable. This significantly simplifies the computation of the posterior and makes SVGD a powerful tool for inference of intractable distributions \cite{steinNIPS2016_6338} as usually encountered in nonlinear filtering. The SVGD algorithm follows from a closed form solution of \eqref{eq:phi_star}, as shown in \cite{steinNIPS2016_6338,ksd110.5555/3045390.3045665,ksd210.5555/3045390.3045421}, and is given by:
\begin{equation}
\boldsymbol{\phi}^*(\cdot)  = \mathbb E_{\mathbf{x}\sim q}[ \nabla_{\mathbf{x}} \log p(\mathbf{x}) k(\mathbf{x},\cdot) + \nabla_{\mathbf{x}} k(\mathbf{x},\cdot)].
\label{equ:phistar_closed_form}             
\end{equation}
Equation \eqref{equ:phistar_closed_form} provides the optimal update direction for the particles within $\mathcal H_k^d$.
In practice, the set of particles ${\{\mathbf{x}^i\}}_{i=1}^N$ is initialized randomly according to some prior distribution representing $q$. These particles are updated with the approximate steepest direction $\hat {\boldsymbol{\phi}}^* (\mathbf{x}) $ given by:
\vskip-3mm
\begin{equation}
\hat{\boldsymbol{\phi}}^*(\mathbf{x}) =\frac{1}{N}\sum_{j=1}^N [\nabla \log p(\mathbf{x}^j)k(\mathbf{x}^j, \mathbf{x}) +  \nabla_{\mathbf{x}^j}k(\mathbf{x}^j,\mathbf{x})], 
\label{stein_phi_approx}
\end{equation}
\vskip-0.1cm
\noindent where $k(\mathbf{x},\mathbf{x}')$ is a positive definite kernel. 

The first component in \eqref{stein_phi_approx} can be interpreted as a weighted gradient of the log posterior with weights given by the kernel function evaluated on the set of particles. This pushes the particles towards the modes of the posterior. The second component is the gradient of the kernel at the particle locations and corresponds to a repulsive force bringing diversity among the particles. If a particle is within the vicinity of another particle, the second term will tend to separate them, thus preventing them from collapsing into a single point. These two components balance each other so that the resulting particle set can approximate complex multi-modal posterior distributions.

SVGD uses first order gradients to sample the posterior distribution and solves the optimization problem in \eqref{stein_phi_approx} using mini-batch stochastic gradient descent (SGD). 
For non-convex problems, however, gradients based methods can be slow in convergence~\cite{var_red}. This is the motivation to include second order curvature information in our proposed method which we describe in the next section.

% +-----------------------------------------------------------------------------
% | Stein Particle Filter
% +-----------------------------------------------------------------------------
\section{Stein Particle Filter}
\label{sec:spf}

In this section we describe the prediction and update steps of our filtering technique. As both accuracy and speed are critical in filtering problems, we leverage second-order (curvature) information and propose a Stein Quasi-Newton Gradient Descent algorithm based on L-BFGS~\cite{Nocedal2006Numerical}. 

\subsection{Prediction Step}
We first describe how the filtering equations are solved as part of our framework. In the prediction step we marginalize over the previous step's posterior multiplied by the transition model. As with particle filters, the previous posterior is represented by a set of particles $\{\mathbf{x}_{t-1}^j\}_{j=1}^N$ which allows us to simply apply the transition function to propagate the particles to obtain the predictive distribution at time $t$:

\vskip-6mm
\begin{equation}
    p(\mathbf{x}_t|\mathbf{z}_{1:t-1}, \mathbf{u}_{1:t}) \approx \frac{1}{N}\sum_{j=1}^{N} p(\mathbf{x}_t|\mathbf{x}_{t-1}^j,\mathbf{u}_t).
\label{eq:svqn_prediction}
\end{equation}
%\vskip-6mm
\subsection{Update Step}
In the update step, we update the current belief with new observations as per Eq.~\ref{eq:corrected_belief}. The logarithm of the posterior is given by,
\begin{equation}
\log p(\mathbf{x}_t|\mathbf{z}_{1:t}) = \log{\eta} +\log p(\mathbf{z}_t|\mathbf{x}_t) + \; \log p(\mathbf{x}_t|\mathbf{z}_{1:t-1}, \mathbf{u}_{1:t}),
\label{eq:posterior}
\end{equation}
which is used in Eq.~\ref{stein_phi_approx} 
to propagate the particles in Eq.~\eqref{eq:qn_update} and integrate the new sensor observation. We run a few iterations of Eq.~\ref{eq:qn_update} 
(typically between 10 and 50) to converge to an accurate posterior. Note that $\log \eta$ does not depend on $\mathbf{x}$ hence its derivative is zero and does not incur in extra computational cost. Finally, we can rewrite the posterior expression as,
\vskip-5mm
\begin{equation}
    p(\mathbf{x}_t|\mathbf{z}_{1:t}, \mathbf{u}_{1:t}) \propto p(\mathbf{z}_t|\mathbf{x}_t) p(\mathbf{x}_t|\mathbf{z}_{1:t-1},\mathbf{u}_{1:t}),
    \label{eq:svqn_update}
\end{equation}
where $p(\mathbf{x}_t|\mathbf{z}_{1:t-1},\mathbf{u}_{1:t})$ is the predictive distribution given by Eq.\ref{eq:svqn_prediction}.

\subsection{Stein Quasi-Newton Gradient Descent}

{\blue We incorporate second-order information into the standard SVGD algorithm in two ways, in kernel scaling and gradient flow, without a substantial change in its main properties. First, we use curvature information represented by the Hessian of the logarithm of the target density to specify an anisotropic kernel that better captures the geometry of the target density.} This idea has been used in the Stein variational newton (SVN) method~\cite{detommaso_2018} within a full Newton extension of SVGD. The Hessian scaled RBF kernel used in our method is defined as,
$k(\mathbf{x},\mathbf{x}') = \exp\left(- \frac{1}{d}(\mathbf{x}-\mathbf{x}')^\top M (\mathbf{x}-\mathbf{x}')\right), 
$ where $d$ is the dimensionality of $\mathbf{x}$ and $M$ {\blue approximates the expected curvature. }
Using $A(\mathbf{x})$ to denote the local approximation of the Hessian of the negative log-target density at a particle location, $A(\mathbf{x}) \approx - \nabla_{\mathbf{x}}^2 \log p(\mathbf{x})$, we can define $M := \frac{1}{N} \sum_{i=1}^N A(\mathbf{x}^i)$.

The effect of using curvature information to compute the kernel is to deform the space in the directions of higher variations, making the particles flow more evenly to better capture higher probability regions.

We also scale the gradient update in Eq.~\ref{eq:phi_star_up} by a {\blue positive definite} pre-conditioner derived from the Hessian. {\blue This accelerates the convergence rate in the direction of the curvature}. As for high-dimensional problems, the Hessian can be very expensive to compute we adopt a quasi-Newton solution based on L-BFGS that iteratively approximates the inverse Hessian as,
\vskip-6mm
\begin{align}
    H_{k+1} & = (I - \rho_k \mathbf{s}_k \mathbf{y}_k^T) H_k (I - \rho_k \mathbf{y}_k \mathbf{s}_k^T) + \rho_k \mathbf{s}_k \mathbf{s}_k^T \\
    \mathbf{s}_k & = \mathbf{x}_{k+1} - \mathbf{x}_k \\
    \mathbf{y}_k & = \nabla_{\mathbf{x}_{k+1}} \log p(\mathbf{x}_{k+1}) - \nabla_{\mathbf{x}_{k}} \log p(\mathbf{x}_{k}).
\end{align}
\vskip-0.2cm
In the above, $I$ is the identity matrix, $\rho_k=\frac{1}{\mathbf{y}_k^T \mathbf{s}_k}$, and the initial solution is usually set to be a diagonal approximation to the inverse Hessian. The updated equation for the Stein particle flow is then, 
\vskip-0.7cm
\begin{align}
    &\mathbf{x}^j =
   % \gets 
    \mathbf{x}^j  + \epsilon H \hat{\boldsymbol{\phi}}^*(\mathbf{x}^j).
   \label{eq:qn_update}
\end{align}
\vskip-0.3cm

Note that to update the prediction particles, 
Stein particle filter (SPF) uses a quasi-Newton approximation to the inverse Hessian of the cost function which acts as a pre-conditioner to the gradient update. The approximated Hessian incorporates the history of gradients which accounts for the curvature of the log posterior. This significantly improves the convergence rate of the method compared to standard SVGD. The particle update is performed in $L$ iterations of the L-BFGS algorithm which allows the estimate of the Hessian of the observation likelihood to iteratively correct the prediction distribution based on new observations. 
Similar to particle filters, particles are initialized uniformally for global state estimation problems. Then for each next time step, particles utilize the previous updated state as a prior distribution. In contrast to particle filters, SPF estimates the posterior distribution with equal weight particles and does not require a resampling step, thus eliminating the potential particle impoverishment problem commonly observed in PFs. The core steps of SPF are outlined in Algorithm~\ref{alg:spf_alg}.
\begin{algorithm}[bt]
	\caption  {\blue One step of Stein particle filter}
	\label{alg:spf_alg}
	
	\SetAlgoLined
  	\DontPrintSemicolon
    \SetKwInOut{Input}{Input}
    %\SetKwInOut{Output}{Output} 
    \Input{$X_{t-1}$, $\mathbf{u}_t$, $\mathbf{z}_t$
        \\
    }
    
	$\forall_{j=1:N} \; \; \mathbf{x}_t^j \sim  \;p(\mathbf{x}_t^j|\mathbf{z}_{1:t-1}^j,\mathbf{u}_{1:t})$ \tcp*{Prediction step using \eqref{eq:svqn_prediction}}
		\For {$l=1,2,\dots,L$}
		{
		    $\mathbf x^j_{l+1} \gets \mathbf x^j_l + \epsilon H_l \boldsymbol{\hat\phi^*}(\mathbf x^j_l) \quad \forall j=1,\dots,N$ \tcp*{Update step with L-BFGS \eqref{eq:qn_update}}
		}
    \Return {$X_{t}$}\;
    \vspace{-1mm}
\end{algorithm}

% +-----------------------------------------------------------------------------
% | Experiments
% +-----------------------------------------------------------------------------
\section{Experiments}

In the experiments we demonstrate the ability of the proposed SPF method to provide accurate state estimation while requiring significantly fewer particles compared to a particle filter (PF) {\blue which uses low variance sampling}{\blue ~\cite{probabilistic_robotics}}. In Section \ref{multi-tr}, we first demonstrate the efficiency and accuracy of our proposed method on a synthetic task and provide comparisons to a PF. 
{\blue To showcase the improved convergence rate of our method and the relative benefit of adding second-order information and L-BFGS compared to other gradient-based methods, 

we employ SVGD~\cite{steinNIPS2016_6338} and SVN~\cite{detommaso_2018} within the particle filter framework to obtain SVGDPF and SVNPF respectively and use Adam~\cite{adam} as their optimizer. SVN accelerates the convergence of the SVGD algorithm by exploiting the second-order information 
in Stein variational framework.  
SVGDPF is our 
MPF~\cite{mapping_particle_f} implementation and can be seen as a particular case of SPF, i.e., SVGDPF is SPF with no second order information and Adam as the optimizer.
} 
In Section \ref{high_dim}, we demonstrate the ability of SPF to scale to higher dimensional problems with limited particle count. Finally, in  Section \ref{likelihood_grad_explained}, we  evaluate our method in a challenging 3D localization task.
{\blue All experiments were performed on a desktop PC with an Intel Core i7-7700 CPU and 16 GB RAM.}
\begin{figure}[!bth]
\centering
\begin{minipage}{0.35\textwidth}

\includegraphics[width=0.98\textwidth,height=0.28\textwidth]{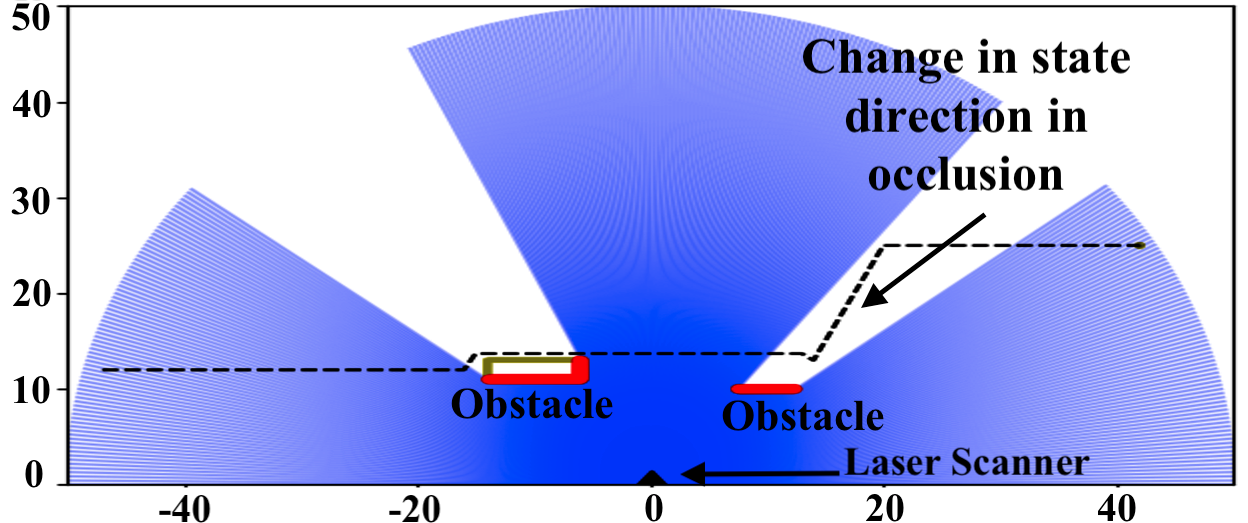} \\[1ex]
\vspace{-0.7cm}
\subcaption[]{}
\label{fig:bi-modal-a}

\includegraphics[width=0.98\textwidth]{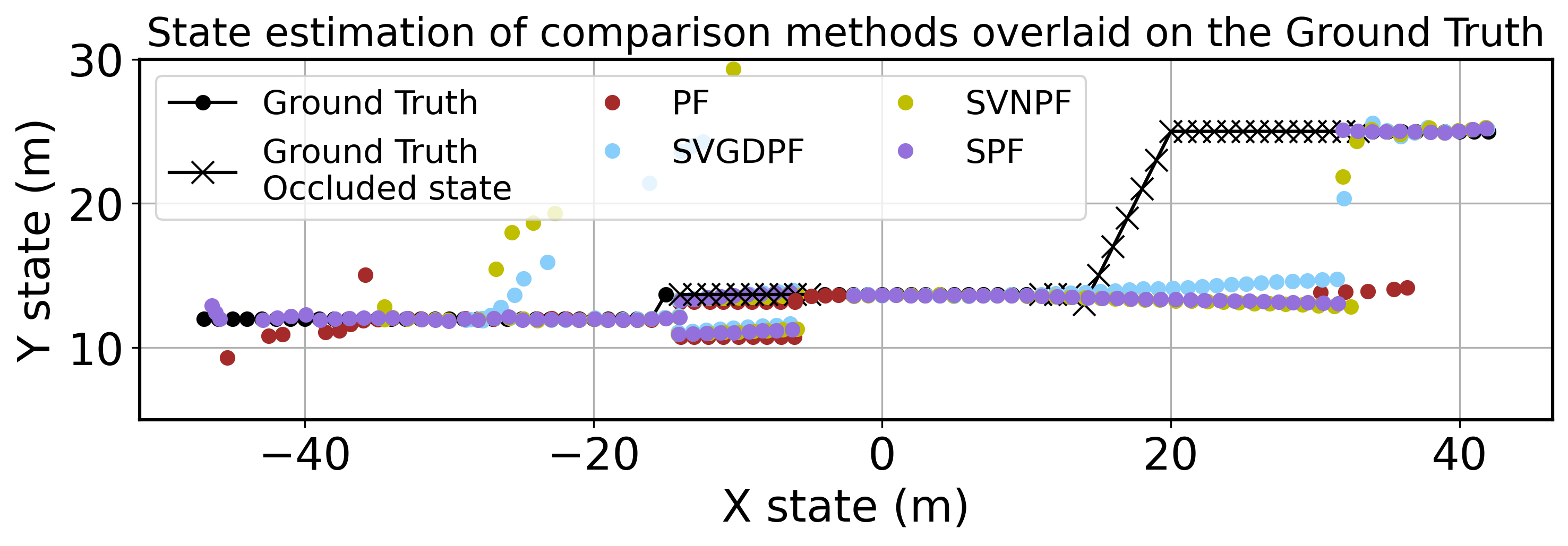} \\[1ex]
\vspace{-0.7cm}
\subcaption[]{}
\label{fig:bi-modal-b}

\includegraphics[width=0.98\textwidth]{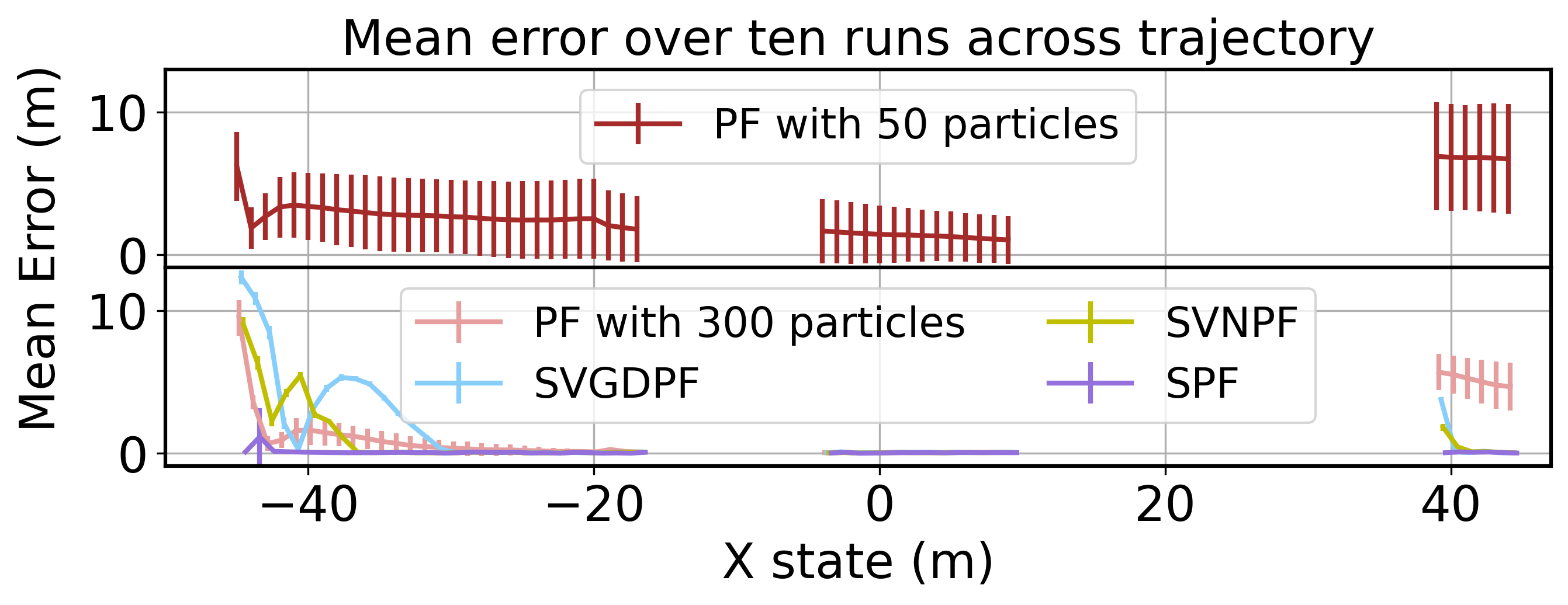} \\[1ex]
\vspace{-0.7cm}
\subcaption[]{}
\label{fig:bi-modal-c}
\end{minipage}
\hfill
\begin{minipage}{0.12\textwidth}
\includegraphics[width=0.63\textwidth]{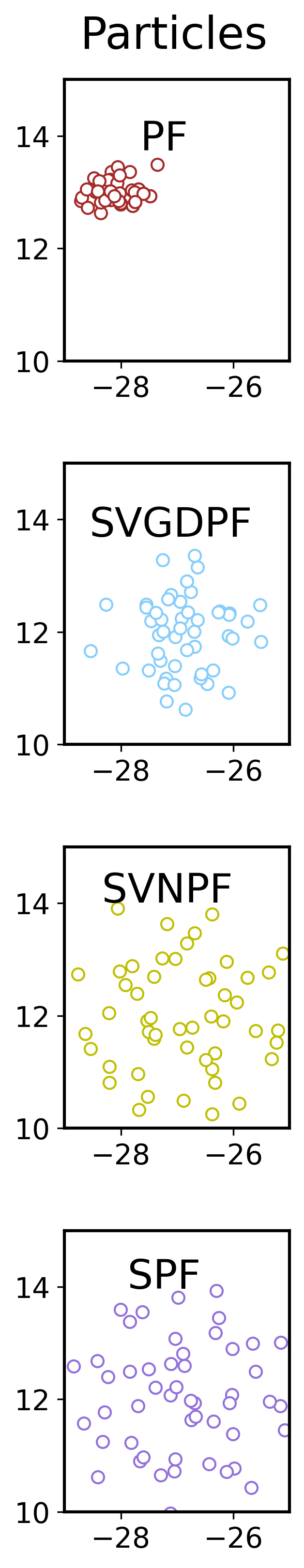} 
 \subcaption[]{}
\label{fig:bi-modal-d}
\end{minipage}
\captionsetup{singlelinecheck = false, justification=justified}
\setlength{\belowcaptionskip}{-13pt}
\caption{State estimation in a simulated 2D environment with $50$ particles. (a) simulated environment with two obstacles. 
(b) Trajectory estimates overlaid on the ground truth. When robot looses the true trajectory in occlusion, only gradient based methods recover the true state immediately upon receiving the laser readings, while PF continues on in the wrong direction. (c) Corresponding mean error across different parts of the trajectory showing earlier convergence of SPF. Top subplot shows poor performance of PF with $50$ particles. Bottom subplot shows improved performance of PF with $300$ particles. PF still requires few time steps to converge owing to it's overoptimistic particle distribution shown in { \ref{fig:bi-modal-d}.}}
\end{figure}

\begin{figure}[bth]
    \centering
    \includegraphics[page=1,width=0.48\textwidth]{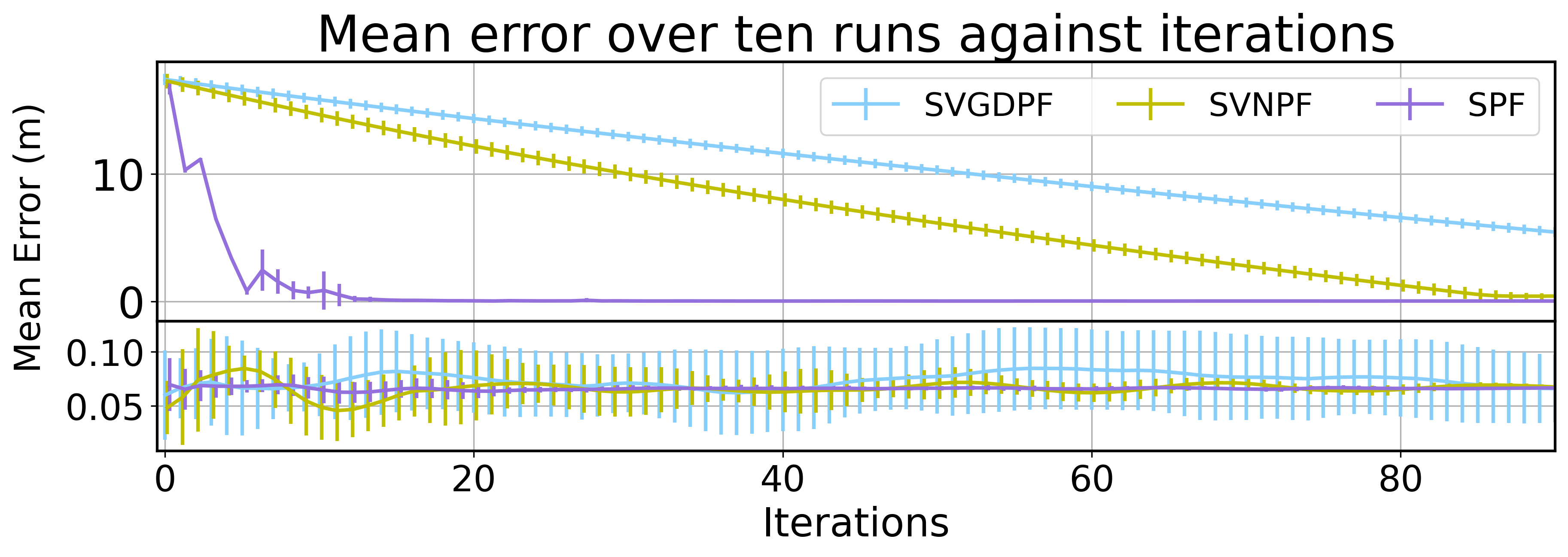}
    \vspace{-6mm}
    \setlength{\belowcaptionskip}{-15pt}
    \caption{Mean error of gradient-based methods against number of iterations for the first time step (top) and $15^{th}$ time step ({\blue when all methods have converged to the true state}) (bottom). SPF converges earlier to the correct state 
    while other methods take much longer. Bottom subplot shows the performance in tracking mode indicating instant convergence requiring a few iterations.}
   \label{fig:error_versus_iter} 
   
\end{figure}
\subsection{Multi-modal Tracking}\label{multi-tr}
In this experiment we demonstrate the capability of SPF to track the robot accurately in multi-modal occluded scenes and its ability to recover the correct mode upon receiving new observations. Through this experiment we also validate the better convergence rates of SPF in comparison to SVGDPF and SVNPF. In this experiment a simulated moving robot 
needs to be tracked over time. The scene in Fig.~\ref{fig:bi-modal-a} is observed by a static laser scanner which provides observations to the filter to track the state of the robot. As the robot reaches the first obstacle it follows one of two possible paths. Upon reaching the second obstacle to the laser scanner, robot changes it's heading in occlusion. 
To track the state of the moving robot, all methods use a simple constant velocity motion model{\blue ~\cite{probabilistic_robotics}}. 
The likelihood function of the update model in this experiment is the Euclidean distance between the laser measurements and the proposal distribution{\blue~\cite{probabilistic_robotics}}. The gradients are then obtained from this optimization problem by minimizing the distance between the state obtained with the observation and the proposal distribution. Gradient-based methods use these gradients to update the state of the robot at the update step using a fixed number of iterations, $35$ in this instance. Each method is run $10$ times using $50$ particles and the error bars over these runs are shown across the entire trajectory in Fig.~\ref{fig:bi-modal-c}. Blank spaces represent the occluded motion where we do not compute the errors for clarity.

Fig.~\ref{fig:bi-modal-b} shows the quality of estimated trajectories overlaid on the ground truth trajectory and Fig.~\ref{fig:bi-modal-d} shows the corresponding particle distributions at a specific time step {\blue after all methods have converged}. Fig.~\ref{fig:bi-modal-d} shows that the particle spread with PF state estimation is more condensed, which is due to the resampling of  particles, compared to gradient-based methods where the repulsive force among the particles induces a reasonable spread. The PF particles represent an over-optimistic estimation of the posterior which tends to cause slow convergence to the true state distribution even with $300$ particles as shown at the bottom subplot of~\ref{fig:bi-modal-c}.

Fig.~\ref{fig:bi-modal-c} shows an earlier convergence of SPF to the true state, both in the beginning and towards the end when filter receives observation after robot changes its heading in occlusion, compared to other methods when run for a fixed number of iterations as indicated by the lowest error from the very initial time steps. The approximate curvature information exploited by SPF when optimizing the objective function scales the gradient direction by constructing a positive definite matrix resulting in faster convergence. SVNPF also uses second order information in constructing the steepest direction and achieves earlier convergence compared to SVGDPF. 
The convergence of PF purely relies on the chance of particles being close to the true state in the global initialization after which it requires a few time steps to converge. PF performs poorly with $50$ particles in this environment as shown in Fig.~\ref{fig:bi-modal-c} (top) and achieves roughly equivalent performance to that of gradient based methods with $300$ particles shown in the Fig.~\ref{fig:bi-modal-c} (bottom). 

Figs.~\ref{fig:bi-modal-b} and ~\ref{fig:bi-modal-c} also demonstrate the ability of gradient based methods to recover the true state, 
after a long sequence of update steps with no observations due to the change in the robot's heading while occluded by an obstacle.
In these type of occluded scenarios, a filter typically predicts the motion in a straight line following a constant velocity motion model. 
Once laser measurements become available only gradient-based methods are able to recover the true trajectory while PF is not, as indicated by the corresponding large error in Fig.~\ref{fig:bi-modal-c}. 

Fig.~\ref{fig:error_versus_iter} presents detailed convergence analysis of gradient-based methods over number of iterations by recording the error over $10$ evaluations, each run for $90$ iterations. The recordings are shown for the first (top) and the fifteenth (bottom) time step to highlight the difference in convergence behaviour for the global state estimation and tracking tasks respectively. This figure shows that all gradient based methods take considerably longer time to perform global state estimation, as indicated by large errors before convergence (top), compared to performing the tracking tasks (bottom) where a few iterations suffice. However, among these methods, SPF, again, achieves much earlier convergence, followed by SVNPF. 
This experiment validates the benefits of exploiting curvature information in the estimation process compared to first order gradient information. 
\subsection{Solution Quality with limited particle count}\label{high_dim}
In this experiment, we showcase the SPF's ability to scale to higher dimensions while being limited in particle count  
where PF does not work. The task is to track the phase shift and amplitude of a sin wave of the form $g(t)=A \sin({k(t+\phi)})$, where $A$, $k$, and $\frac{\phi}{k}$ are amplitude, period, and horizontal phase shift respectively. 

We start with two sin functions and increase the dimensionality up-to $10$ functions using $50$ particles and compare the quality of SPF and PF estimates. Each sin function is represented by a two dimensional state space comprised of phase shift and amplitude giving us $20$ dimensional state space for $10$ sin functions. The filters receive the noisy observations of the true function value which are used in the update model in constructing the likelihood function as a Euclidean distance between the prediction distribution and the observations. SPF computes the gradients of amplitude and phase shift from the likelihood function to update the SPF estimate of the function. Table~\ref{tab:dime} presents the comparison of SPF and PF in estimating the sin wave function over time using root mean squared error (RMSE). Table~\ref{tab:dime} demonstrates the ability of SPF to scale to $20$ dimensions with just $50$ particles.

This experiment shows that SPF can scale to higher dimensions with a limited number of particles by exploiting gradients of the likelihood function. Using the gradient information allows the update model to correct the predicted state, thus improving the quality of state estimate. In contrast, the computational complexity of PF, in terms of required number of particles, increases with the increase in the state dimension. As shown by high RMSE in the Table~\ref{tab:dime} PF is not able to estimate the state using $50$ particles and requires thousands of particles to estimate the state reasonably. The last column of the table shows 
the improved state estimation of PF with $10000$ particles.

\begin{table}[bht]
%\blue{
	
		\caption{State estimation errors in high dimensions. }
		\label{tab:dime}
	    \begin{tabular}{lll>{\blue}r%>{\blue}c
	    }
    		\toprule   
    		    
    	    \multirow{2}{*}{\raisebox{-\heavyrulewidth}{\rotatebox[origin=c]{90}
    	    {Dim.}}} & 
    	    \multicolumn{1}{c}{SPF}  &
    	    \multicolumn{1}{c}{PF}  &
    	    \multicolumn{1}{c} {\blue PF with more particles}\\
            \cmidrule{2-4}
&\multicolumn{2}{c}{RMSE with $50$ particles}  & %particles &RMSE   \\
\multicolumn{1}{c}{\blue RMSE with $10k$ particles}\\
\cmidrule{1-4}
4 & $0.82 \pm 0.63$ & $105.05 \pm 47.83$
&$8.29 \pm 6.44$\\ 
8 & $1.04 \pm 0.62$ &$265.63 \pm 62.33$
&$27.03 \pm 11.69$\\ 
12 & $1.39 \pm 1.02$ & $319.71 \pm 42.99$
&$63.08 \pm 29.81$\\
16 & $1.14 \pm 0.69$ &$366.62 \pm 37.28$ 
&$89.65 \pm 27.48$ \\
20 & $1.14 \pm 0.44$ &$407.12 \pm 32.66$
&$123.89 \pm 29.98$ \\

		    \bottomrule 
	    \end{tabular}
%	\end{center}
%\vskip-5mm
%}
\end{table}

\iffalse
\begin{figure}[bt]

    \centering

    \begin{subfigure}{.48\textwidth}
	   
	    \includegraphics[page=0.8,width=0.98\textwidth]{sine/limited2_10_ with newton stein.png}
	    
   \end{subfigure}
   \begin{subfigure}{.48\textwidth}

    \includegraphics[page=0.8,width=0.98\textwidth]{sine/limited2_10_conventional_.png}
%    
   \end{subfigure}
   \begin{subfigure}{.48\textwidth}

      \includegraphics[page=0.8,width=0.98\textwidth]{sine/limited2_10_conventional.png}
%    
   \end{subfigure}
    
    \setlength{\belowcaptionskip}{-18pt}
    \caption{
    Each plot of a sin curve shows the results for five dimensions for the $10$ sin functions. (Top): Quality of SPF's estimated trajectory using $50$ particles. (Bottom): PF fails to estimate the correct trajectory using $50$ particles. (Bottom): PF's improved trajectory using $10000$ particles.
    }
   
  \label{fig:scalability_fig}  	
\end{figure}
\fi
\subsection{Localization Task}\label{likelihood_grad_explained}
In this experiment we showcase the particle count efficiency of SPF in two case studies: i) a global localization task and ii) a tracking task, both using the 3D LiDAR data of the Newer College dataset \cite{newcollege}.
We compare the localization accuracy of SPF and PF using the root mean squared error (RMSE) in Section~\ref{track} and present a run-time comparison in Section~\ref{time-stud}. {\blue The localization experiments were performed with a C++ implementation}. 
In the prediction step, both methods propagate the particles towards the proposal distribution using an SGDICP-based~\cite{fahira2018} motion model. In the update step, each particle uses an observation likelihood model of the following form:
$\forall_j \; \;   p(\mathbf{z}_t| \boldsymbol{x}_t^j, \ \mathrm{map}) = \frac{1}{K}\sum_{i=1}^K \frac{d_{i^j}^2}{\sigma^2} $, where the $\mathrm{map}$ is represented by an octomap~\cite{octomap} with a \SI{0.2}{\meter} resolution, $K$ is the total number of beams in the point cloud, $\sigma$ is the standard deviation of the distance measurements of a single beam. 
Finally,
\begin{equation}
d_{i^j}^2=||T_{\mathbf{x}^j} \mathbf{b}_i-\mathbf{y}||^2 = ||T_{\mathbf{x}^j} \mathbf{b}_i-\textit{NN}(T_{\mathbf{x}^j}\mathbf{b}_i)||^2,
\label{eq:sdf}
\end{equation}
is the Euclidean distance between the end of the $i^{th}$ beam $\mathbf{b}_i\in \mathbb R^3$ and its nearest neighbor $y = NN(T_{\mathbf{x}^j}\mathbf{b}_i) \in \mathbb R^3$, when projected into the map using the particle's predicted state $\mathbf{x}^j$ as transformation $T_{\mathbf{x}^j} \in \mathbb R^{4 \times 4}$. This captures how well the particle's pose estimate explains the observations of the environment and ideally is zero. To compute the gradients, the rigid body transformation $T_{\mathbf{x}^j}$ is decomposed into six terms corresponding to the three translation and three rotation components of the state of the robot.

\subsubsection{Global Localization}
We constrain the 6D localization problem to a 4D one by limiting roll and pitch within $2^\circ$ and perform the global localization using just $50$ particles. Particles are spread uniformly over the entire map with the elevation $z$ being restricted to \SI{10}{m} height. This is done to confine the particles to the map boundaries even in areas with vegetation which lack tall wall structures. To ensure that particles are distributed uniformly, yaw values are constrained to increments of $30^\circ$. These steps ensure that particles roughly cover the entire state space while being limited in number.

Non-convex error landscapes with multiple local minimas can potentially slow down the SPF's convergence to the true state. In order to obtain a single or a few likely modes without performing a resampling step, we propose a re-projection step which guides particles in unlikely locations towards states of more likely particles. To this end we compute a new gradient by using the state of the more likely particle $x_k$ in the nearest neighbor search of \eqref{eq:sdf}. This results in a gradient that pulls the particle towards $x_k$ while still allowing variation in the convergence. To ensure that the observation model can find a solution we make use of a representation proposed in~\cite{global_icp} which augments the 3D points of the scan with a fourth dimension that represents the Euclidean distance of each point to the centroid of the scan. This rotational invariant information aids in finding good point correspondences.

Allowing particles to follow the new gradient direction gives them a chance to converge to a state closer to the more likely particles' state while being able to choose different state during the particle interaction in the RKHS. This is in contrast to performing a resampling step which replaces the less likely particles with the more likely ones. 

Particles for which no observation exist in the map at a given state, acquire the matched observation pairs of other particles which have a larger number of matching point pairs. This can happen when a particle is propagated outside the map boundaries at which stage the observation model can no longer match to the closest observation in the map.

Once particles are initialized uniformly, SPF is run for a few time steps to allow the particles to converge to possibly several local modes.
Ambiguous or noisy observations

can result in multiple-hypothesis during the localization process which are corrected using the re-projection step.

\begin{figure}[bt]
\centering
\begin{minipage}{0.48\textwidth}
\vspace{0pt}
 
\includegraphics[width=\textwidth]{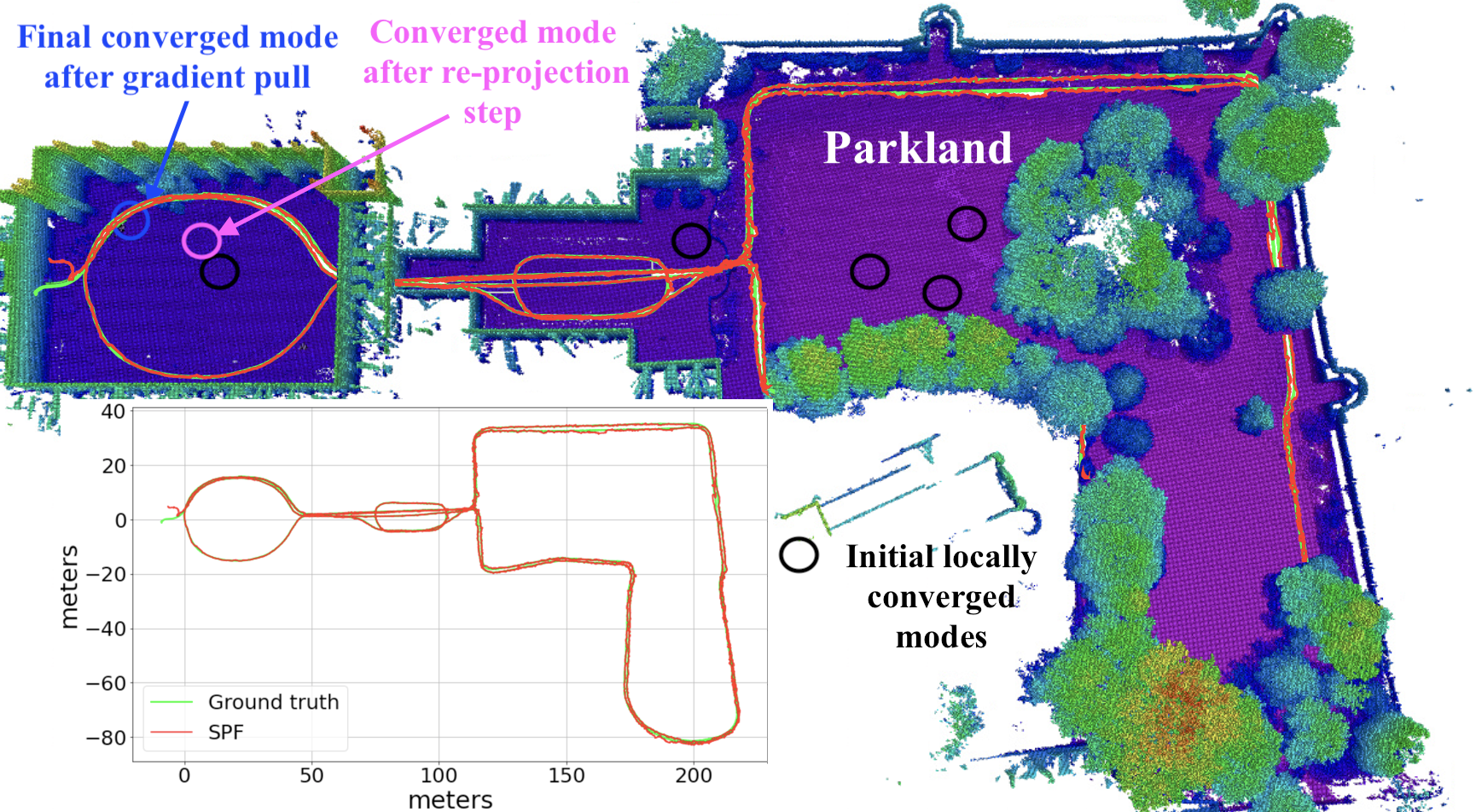} 
\vspace{1 mm}
%\subcaption[]{}
\end{minipage}

\vspace{-0.7cm}
\setlength{\belowcaptionskip}{-8pt}
\caption{ 
Global localization results of SPF. Re-projection step brings the locally converged particles closer to the true closest obstacle in the map where gradients correct their state in the update step. SPF trajectory is overlaid on the ground truth showing high quality of the estimated states.}
\label{fig:global_loc}
\end{figure}

Fig.~\ref{fig:global_loc} shows the initial local modes SPF recovers in the global localization task. We can see that the SPF converged mode after re-projection step (shown in pink circle) is far from the true state. Since SPF corrects the predicted state during the update step by exploiting the gradients of the observation likelihood model, these particles ultimately converge to the correct mode as shown in blue circle. 
When SPF converges multiple modes, only one of them survives over the time. Others either die out by crossing the map boundaries or converging to the correct mode. In this task, PF, with $50$ particles, in all $20$ runs converged to a wrong state at different spots in the Parkland. In this experiment PF requires roughly $1000$ particles to reliably converge to the correct mode. SPF with $50$ particles and PF with $1000$ particles successfully converge to the true state for $9$ and $6$ times respectively out of $10$ runs highlighting the improved performance of SPF even with significantly fewer particles.
This experiment confirms the benefit of avoiding the resampling step to overcome the particle impoverishment problem and highlights the particle efficiency of SPF in comparison to PF.

\begin{table}[bt]
	\begin{center}
		%\vspace{-5mm}
		\caption{Localization Errors (m) with $10^{th}$ and $90^{th}$ quantiles in a tracking setup %for SPF and PF 
		with 5, 20, and 50 particles.}
		\label{tab:loc_accuracy}
	  
	    \begin{tabular}{llrrr}
    		\toprule

Methd.&&5 &20  &50 \\
%& 5  &20  &50 
%& 5  &20  &50 \\
\cmidrule{1-5}
%\cmidrule{1-10}
SPF &RMSE& $ 1.19 \pm 0.74 $& $ 1.02 \pm 0.64    $ & $ 0.62 \pm 0.37 $
\\&Quantiles& $  0.31 , 1.76 $ & $  0.28 , 1.41  $ &$  0.18 , 0.88  $ \\

PF &RMSE&$106.19 \pm 73.24 $ & $ 1.75 \pm 1.36  $ & $ 0.72 \pm 0.51 $
\\&Quantiles& $  1.38 , 194.54  $ & $  0.21 , 2.82 $&$  0.17 , 0.91  $
%\\&Time (ms)& $  1.38 , 194.54  $ & $  0.21 , 2.82 $&$  0.17 , 0.91  $

\\ \vspace{-0.75em} \\

		    \bottomrule 
	    \end{tabular}
	\end{center}
	\vspace{-5mm}
	\vskip-3mm
\end{table}

\subsubsection{Tracking}\label{track}

In this task both filters start with the correct state to track the Newer College dataset trajectory using $5$, $20$, and $50$ particles. Table~\ref{tab:loc_accuracy} shows the RMSE of the localization error for both SPF and PF in meters, both mean and standard deviation as well as the $10\%$ and $90\%$ quantiles are reported. While for both 20 and 50 particles both methods achieve comparable results, SPF always achieves lower variance results. When using only 5 particles the PF diverges where as the SPF still achieves acceptable results. This showcases how even in scenarios where a PF works as expected there is benefits in robustness and number of particles required for our proposed SPF.
\subsubsection{Run-Time Analysis}\label{time-stud}
Compared to PF, for each time step SPF incurs an extra cost of $10$ to $20$ iterations for tracking and $10$ to $60$ iterations for global localization task. The PF's and SPF's run-time for %each time 
update step with fixed $20$ iterations and the extra per iteration cost of SPF is shown in Table~\ref{tab:runtime}. This table shows that SPF bears only minor overhead of few milli-seconds for each iteration resulting in { \blue a computationally efficient filtering method.} 
\begin{table}[bt]
	\begin{center}
		
		\caption{Run-time (milliseconds) 
		with 5, 20, and 50 particles.}
		\label{tab:runtime}
	    \begin{tabular}{llrr}
    		\toprule

    	    Method

&5 &20  &50 \\
\cmidrule{1-4}

PF& $40.28 \pm 10.03$& $82.40 \pm 15.13$ & $108.20 \pm 29.85$
\\
SPF&$42.88 \pm 10.04$ & $89.80 \pm 15.20$ & $143 \pm 30.14$
\\
\multirow{1.5}{*}{\small{SPF's extra}} &\multirow{1.5}{*}{$0.13 \pm 0.01$} & \multirow{1.5}{*}{$0.37 \pm 0.7$} &\multirow{1.5}{*}{$1.74 \pm 0.29$}
\\ cost/iteration
\\ \vspace{-0.75em} \\

		    \bottomrule 
	    \end{tabular}
	\end{center}
	\vspace{-8mm}
\end{table}

% +-----------------------------------------------------------------------------
% | Conclusion
% +-----------------------------------------------------------------------------
\section{Conclusion}
In this paper we introduced a novel particle filter which exploits the Stein variational framework. Our proposed filter transports proposal particles to the target distribution using gradient information. 
{\blue This flow of particles is embedded in a reproducing kernel Hilbert space where particles’ interaction is taken into account to bring diversity among the particles, and moving them harmoniously even when in areas with low or zero probability mass.} As a result our proposed method requires fewer particles to approximate the posterior while being able to scale to the higher dimensions. 
Our method does not require the resampling step which tends to lead to particle impoverishment problem. Experiments on simulated as well as real datasets demonstrate that our method is more particle efficient as well as able to recover the posterior distribution in multi-modal occluded scenes.

{\small
\bibliographystyle{IEEEtran}
\bibliography{paper}  % .bib

\begin{thebibliography}{10}
\providecommand{\url}[1]{#1}
\csname url@rmstyle\endcsname
\providecommand{\newblock}{\relax}
\providecommand{\bibinfo}[2]{#2}
\providecommand\BIBentrySTDinterwordspacing{\spaceskip=0pt\relax}
\providecommand\BIBentryALTinterwordstretchfactor{4}
\providecommand\BIBentryALTinterwordspacing{\spaceskip=\fontdimen2\font plus
\BIBentryALTinterwordstretchfactor\fontdimen3\font minus
  \fontdimen4\font\relax}
\providecommand\BIBforeignlanguage[2]{{%
\expandafter\ifx\csname l@#1\endcsname\relax
\typeout{** WARNING: IEEEtran.bst: No hyphenation pattern has been}%
\typeout{** loaded for the language `#1'. Using the pattern for}%
\typeout{** the default language instead.}%
\else
\language=\csname l@#1\endcsname
\fi
#2}}

\bibitem{kalman_1960}
R.~E. Kalman, ``{A New Approach to Linear Filtering and Prediction Problems},''
  \emph{Journal of Basic Engineering}, 1960.

\bibitem{ekf_re}
A.~Gelb and T.~A.~S. Corporation, \emph{Applied Optimal Estimation}.\hskip 1em
  plus 0.5em minus 0.4em\relax The MIT Press, 1974.

\bibitem{doucet_2001_introduction}
A.~Doucet, N.~De~Freitas, and N.~Gordon, ``An introduction to sequential monte
  carlo methods,'' in \emph{Sequential Monte Carlo methods in practice}.\hskip
  1em plus 0.5em minus 0.4em\relax Springer, 2001.

\bibitem{steinNIPS2016_6338}
Q.~Liu and D.~Wang, ``Stein variational gradient descent: A general purpose
  bayesian inference algorithm,'' in \emph{NIPS}, 2016.

\bibitem{Nocedal2006Numerical}
J.~Nocedal and S.~J. Wright, \emph{{Numerical Optimization}}, 2nd~ed., ser.
  Springer Series in Operations Research and Financial Engineering, New York,
  2006.

\bibitem{ukf_rebut}
S.~J. Julier and J.~K. Uhlmann, ``New extension of the kalman filter to
  nonlinear systems,'' in \emph{Defense, Security, and Sensing}, 1997.

\bibitem{weiss_2012}
S.~Weiss, M.~Achtelik, S.~Lynen, M.~Chli, and R.~Siegwart, ``Real-time onboard
  visual-inertial state estimation and self-calibration of mavs in unknown
  environments,'' in \emph{ICRA}, 2012.

\bibitem{thrun_2006_stanley}
S.~Thrun, M.~Montemerlo, H.~Dahlkamp, D.~Stavens, A.~Aron, J.~Diebel, P.~Fong,
  J.~Gale, M.~Halpenny, G.~Hoffmann, K.~Lau, C.~Oakley, M.~Palatucci, V.~Pratt,
  and P.~Stang, ``Stanley: The robot that won the darpa grand challenge,''
  \emph{Journal of Field Robotics}, 2006.

\bibitem{grip_2019}
H.~Grip, J.~Lam, D.~Bayard, D.~Conway, G.~Singh, R.~Brockers, J.~H. Delaune,
  L.~Matthies, C.~Malpica, T.~Brown, A.~Jain, M.~San~Martin, and G.~Merewether,
  ``Flight control system for nasa's mars helicopter,'' in \emph{AIAA Scitech},
  2019.

\bibitem{kothari_2012}
N.~Kothari, B.~Kannan, E.~Glasgwow, and M.~Dias, ``Robust indoor localization
  on a commercial smart phone,'' \emph{Procedia Computer Science}, 2012.

\bibitem{thomas_2007}
U.~Thomas, S.~Molkenstruck, R.~Iser, and F.~Wahl, ``Multi sensor fusion in
  robot assembly using particle filters,'' in \emph{IEEE International
  Conference on Robotics and Automation}, 2007.

\bibitem{duane_1987}
S.~Duane, A.~Kennedy, B.~Pendleton, and D.~Roweth, ``Hybrid monte carlo,''
  \emph{Physics Letters B}, 1987.

\bibitem{choo_2001}
K.~Choo and D.~Fleet, ``People tracking using hybrid monte carlo filtering,''
  in \emph{IEEE ICCV}, 2001.

\bibitem{biswas_2011}
J.~Biswas, B.~Coltin, and M.~Veloso, ``Corrective gradient refinement for
  mobile robot localization,'' in \emph{IROS}, 2011.

\bibitem{rao_black_2001}
A.~Doucet, N.~d. Freitas, K.~P. Murphy, and S.~J. Russell, ``Rao-blackwellised
  particle filtering for dynamic bayesian networks,'' in \emph{Proceedings of
  Conference on Uncertainty in Artificial Intelligence}, San Francisco, CA,
  USA, 2000.

\bibitem{gaussian_pf}
J.~Kotecha and P.~Djuric, ``Gaussian particle filtering,'' \emph{IEEE
  Transactions on Signal Processing}, pp. 2592--2601, 2003.

\bibitem{pf_tut}
M.~Arulampalam, S.~Maskell, N.~Gordon, and T.~Clapp, ``A tutorial on particle
  filters for online nonlinear/non-gaussian bayesian tracking,'' \emph{IEEE
  Transactions on Signal Processing}, vol.~50, pp. 174--188, 2002.

\bibitem{cluster_pf}
A.~Milstein, J.~N. S\'{a}nchez, and E.~T. Williamson, ``Robust global
  localization using clustered particle filtering,'' in \emph{National
  Conference on Artificial Intelligence}, 2002.

\bibitem{kld_sampling}
D.~Fox, ``Kld-sampling: Adaptive particle filters,'' in \emph{NIPS}, 2001.

\bibitem{rebut_uncent_pf}
R.~Merwe, A.~Doucet, N.~Freitas, and E.~Wan, ``The unscented particle filter,''
  \emph{NIPS}, vol.~13, 01 2001.

\bibitem{reset_loc}
S.~Lenser and M.~Eloso, ``Sensor resetting localization for poorly modelled
  mobile robots,'' 2000.

\bibitem{sensor_reset_multi}
B.~Coltin and M.~Veloso, ``Multi-observation sensor resetting localization with
  ambiguous landmarks,'' \emph{Proceedings of the AAAI Conference on Artificial
  Intelligence}, no.~1, Aug. 2011.

\bibitem{mapping_particle_f}
M.~Pulido and P.~J. {van Leeuwen}, ``Sequential monte carlo with kernel
  embedded mappings: The mapping particle filter,'' \emph{Journal of
  Computational Physics}, 2019.

\bibitem{var_red}
M.~Zhu, C.~Liu, and J.~Zhu, ``Variance reduction and quasi-{N}ewton for
  particle-based variational inference,'' in \emph{ICML}, 2020.

\bibitem{gordon_1993}
N.~Gordon, D.~Salmond, and A.~Smith, ``Novel approach to nonlinear/non-gaussian
  bayesian state estimation,'' in \emph{IEEE proceedings of Radar and Signal
  Processing}, 1993.

\bibitem{ksd110.5555/3045390.3045665}
K.~Chwialkowski, H.~Strathmann, and A.~Gretton, ``A kernel test of goodness of
  fit,'' in \emph{ICML}, 2016.

\bibitem{ksd210.5555/3045390.3045421}
Q.~Liu, J.~Lee, and M.~Jordan, ``A kernelized stein discrepancy for
  goodness-of-fit tests,'' in \emph{Proceedings of International Conference on
  Machine Learning}, 2016.

\bibitem{detommaso_2018}
G.~Detommaso, T.~Cui, A.~Spantini, Y.~Marzouk, and R.~Scheichl, ``A stein
  variational newton method,'' in \emph{Proceedings of the International
  Conference on Neural Information Processing Systems}, 2018.

\bibitem{probabilistic_robotics}
S.~Thrun, W.~Burgard, and D.~Fox, \emph{Probabilistic Robotics}.\hskip 1em plus
  0.5em minus 0.4em\relax The MIT Press, 2005.

\bibitem{adam}
D.~P. Kingma and J.~Ba, ``Adam: {A} method for stochastic optimization,'' in
  \emph{International Conference on Learning Representations}, Y.~Bengio and
  Y.~LeCun, Eds., 2015.

\bibitem{newcollege}
M.~{Ramezani}, Y.~{Wang}, M.~{Camurri}, D.~{Wisth}, M.~{Mattamala}, and
  M.~{Fallon}, ``The newer college dataset: Handheld lidar, inertial and vision
  with ground truth,'' in \emph{International Conference on Intelligent Robots
  and Systems}, 2020.

\bibitem{fahira2018}
F.~Afzal~Maken, F.~Ramos, and L.~Ott, ``Speeding up iterative closest point
  using stochastic gradient descent,'' in \emph{IEEE International Conference
  on Robotics and Automation}, 2019.

\bibitem{octomap}
A.~Hornung, K.~M. Wurm, M.~Bennewitz, C.~Stachniss, and W.~Burgard,
  ``{OctoMap}: An efficient probabilistic {3D} mapping framework based on
  octrees,'' \emph{Autonomous Robots}, 2013.

\bibitem{global_icp}
S.~Du, Y.~Xu, T.~Wan, H.~Hu, S.~Zhang, G.~Xu, and X.~Zhang, ``Robust iterative
  closest point algorithm based on global reference point for rotation
  invariant registration,'' \emph{PLOS ONE}, vol.~12, 2017.

\end{thebibliography}

}
\end{document}